\documentclass{article} 
\usepackage{nips13submit_e,times}
\usepackage{hyperref}
\usepackage{url}
\usepackage{graphicx,float}
\usepackage{subfig}
\usepackage{natbib}
\usepackage{amssymb}
\usepackage{amsmath, amsthm}
\usepackage{algorithm}
\usepackage{algorithmic}

\title{Deep Tempering}
\author{
Guillaume Desjardins, Heng G. Luo, Aaron Courville and Yoshua Bengio \\
\{desjagui,heng.luo,courvila,bengioy\}@iro.umontreal.ca \\
D\'{e}partement d'informatique et de recherche op\'{e}rationnelle \\
Universit\'{e} de Montr\'{e}al \\
}

\nipsfinalcopy

\begin{document}

\maketitle

\begin{abstract}
Restricted Boltzmann Machines (RBMs) are one of the fundamental building blocks of
deep learning. Approximate maximum likelihood training of RBMs typically
necessitates sampling from these models. In many training scenarios, computationally efficient Gibbs
sampling procedures are crippled by poor mixing. In this work we propose a novel method of
sampling from Boltzmann machines that demonstrates a computationally
efficient way to promote mixing.
Our approach leverages an under-appreciated property of deep generative
models such as the Deep Belief Network (DBN), where Gibbs sampling from deeper
levels of the latent variable hierarchy results in dramatically increased
ergodicity. Our approach is thus to train an auxiliary latent hierarchical model,
based on the DBN. When used in conjunction with parallel-tempering,
the method is asymptotically guaranteed to simulate samples from the target
RBM. Experimental results confirm the effectiveness of this sampling
strategy in the context of RBM training.
\end{abstract}

\section{Introduction}

Estimating the maximum likelihood gradient of Boltzmann Machines requires
samples to be drawn from the model. This is typically done through Gibbs
sampling where the state of the Markov chain is preserved in between parameter
updates, leading to the Stochastic Maximum Likelihood (SML) algorithm
\citep{Younes1999, Tieleman08-small}. While popular in practice, this algorithm can be
brittle as it relies on the Markov chain mixing properly during training: if
the sampling process fails to adequately explore the space of allowed
configurations (e.g. by getting trapped in regions of high-probability), the
negative phase statistics can become biased and cause the learning procedure to
diverge. Great care must therefore be taken in decreasing the learning rate or
increasing the number of Gibbs steps to offset the loss of ergodicity incurred
by learning.

An attractive and well studied alternative is to augment the Boltzmann
distribution with a temperature parameter in order to perform a simulation in
the joint temperature-configuration space. As the temperature increases, the
distribution becomes more uniform, which in turn allows particles to quickly
explore the energy landscape and escape configurations corresponding to local minima
in the original target distribution. As particles decrease to the nominal
temperature, they are guaranteed to be proper samples of the target
distribution and can thus be used to estimate the sufficient statistics of the
model. \citet{Salakhutdinov-2010, Desjardins+al-2010, Cho10IJCNN,
Salakhutdinov-ICML2010} have all explored variations on this idea, by
considering either serial or parallel implementations of this basic tempering
strategy. This generally results in increased robustness to the choice of
hyper-parameters and faster convergence when used in conjunction with higher
learning rates.

Despite these benefits, tempering methods remain computationally expensive.
While SML requires a mini-batch of $K$ Markov chains to be simulated, Parallel
Tempering (PT) requires $K \times M$ chains, where $M$ is the number
of temperatures. While the serial variants appear to address this issue,
they may be doing so at the expense of sampling efficiency. When using Tempered
Transitions, long-distance jumps can suffer from high-rejection rates,
resulting in the use of possibly stale samples in the gradient estimation step.
Similarly, Coupled Adaptive Simulated Tempering (CAST)
\cite{Salakhutdinov-ICML2010} relies mostly on non-tempered chains to perform
local mode exploration, with non-local moves occurring only occasionally, when
simulated tempering particles are sampled back down to the nominal temperature.
While the relative merits of each method can be discussed at great length, they
unfortunately all share the same Achilles' heel: {\it their tempering parametrization is
inefficient}. They invariably require many interpolating distributions
between the target distribution and the high temperature distribution for
which rapidly mixing sampling is possible.
  
Our proposed solution starts with the idea that we may be able to \emph{learn}
more effective interpolating distributions, allowing us to use far fewer
intermediate distributions than with traditional tempering methods, while still
maintaining a high level of ergodicity in the MCMC simulation of the
extended system. Recent findings \cite{Luo+al-AISTATS2013-small,
Bengio-et-al-ICML2013} have shown that when training deep feature hierarchies,
such as Deep Belief Networks (DBN) \cite{Hinton06} as few as 2-3 layers are
required for the upper layers to mix properly between the modes of the
distribution. It appears as though successive nonlinear transformations to
ever more abstract latent variables (i.e. ever more removed from the data)
play a crucial role in achieving efficient mixing. The goal of this work is
to leverage these learned intermediate distributions to facilitate mixing in
the lower levels of the hierarchy.  

While this procedure is generally applicable to the learning of any
energy-based model featuring latent variables, this paper considers the
particular case of sampling from a Restricted Boltzmann Machine (RBM)
\cite{Smolensky86} for the purpose of training. In this setting, our method can
equivalently be seen as way to perform joint-training of DBNs.

\section{Background}
\label{sec:background}

\paragraph{Restricted Boltzmann Machines.} RBMs define a probability distribution
$p(v,h) = \frac{1}{Z} \exp\left[-E(v,h)\right]$ over a set of visible variables $v \in
\mathbb{R}^{n_1}$ and latent variables $h \in \mathbb{R}^{n_2}$. $E$ is
referred to as the energy function, which in the case of binary visible and
latent units is defined as $E(v,h) = - v^T W h - v^T c -h^T b$. $Z$ is the
partition function defined as $\tiny{Z = \sum_{v,h} \exp\left[-E(v,h)\right]}$. The parameters
$\theta$ of the model include a weight matrix $W\in \mathbb{R}^{n_1 \times
n_2}$ and bias vectors $c \in \mathbb{R}^{n_1}$ and $b \in \mathbb{R}^{n_2}$.
RBMs owe much of their success to this parametrization as it renders its
conditional distributions factorial, enabling trivial inference and efficient
sampling through block Gibbs sampling. This property also allows us to analytically
compute the un-normalized probability of a given configuration $v$ or $h$,
through the free-energy function $F(v),$ such that $p(v) \propto
\exp\left[-F(v)\right]$.

\paragraph{Stochastic Maximum Likelihood.}
The maximum likelihood gradient of an RBM can be derived as the difference of
two expectations,
$\mathbb{E}_q \left[ \frac{\partial F(v)}{\partial \theta} \right] - 
 \mathbb{E}_p \left[ \frac{\partial F(v)}{\partial \theta} \right]$,
where $q(x)$ is the empirical distribution.
Stochastic Maximum Likelihood (SML) \cite{Younes1999,Tieleman08-small}
approximates the expectation under the model distribution (negative phase) through MCMC, the
simplest instantiation being a Gibbs chain which alternates $h \sim p(h\mid
v)$, $v \sim p(v \mid h)$. The particularity of SML is that it avoids the
expensive burn-in process associated with each parameter update by maintaining
the state of the Markov chains in between learning iterations. While asymptotic
convergence of SML has been established under certain conditions\cite{Younes1999},
the method can often diverge in practice when faced with complex multi-modal
distributions. As $p(v)$ becomes increasingly peaked, mode hops can become
exponentially less likely for a Gibbs sampler. No practical learning rate
annealing schedules can then ensure that the model's sufficient statistics are
accurately reflected by the sampler. For this reason, several authors have
proposed using tempering-based samplers to estimate the negative phase
gradient, as these methods are known to be efficient at escaping from regions
of high-probability.

\paragraph{Parallel Tempering.}
We focus here on the Parallel Tempering (PT) strategy of
\cite{Desjardins+al-2010} as it is the most relevant to our work.
Instead of simulating a single Markov chain, PT simulates an ensemble of $M$
models $\prod_{i=1}^M p_i(v)$, with $p_i(v) \propto \exp\left[-\beta_i F(v)
\right]$. The sole difference in the parametrization of $p_i$ wrt. $p:=p_1$ is the
introduction of the inverse temperature parameter $\beta_i \in [0,1]$ which
acts to scale the energy values. Low values of $\beta_i$ act to make the
distributions $p_i$ closer to uniform, and will thus facilitate
sampling.\footnote{Without loss of generality, we define $\beta_0 > \beta_1 >
\cdots > \beta_M$}. To leverage these fast-mixing chains, PT incorporates an
additional transition operator (on top of the standard Gibbs operator
applied independently to each chain): 
swaps (or {\it replica exchanges}) are proposed between samples at neighboring
temperature. If $v_i \sim p_i$, PT accepts this move with probability $r_{i}$
defined as:
\begin{align}
    r_{i}
    &= min\left( 1, \frac
        {p_i\left(v_{i+1}\right) p_{i+1}\left(v_i\right) }
        {p_i\left(v_{i}\right)   p_{i+1}\left(v_{i+1}\right) }
        \right) 
    = min\left( 1, 
        e^{F_i\left(v_{i}\right) + F_{i+1}\left(v_{i+1}\right) -
        F_i\left(v_{i+1}\right) - F_{i+1}\left(v_i\right)}
        \right).
\end{align}
Note that the partition functions are not required to compute this quantity, as
$Z_i$ and $Z_{i+1}$ appear both in the numerator and denominator.  A Gibbs
steps at each Markov chain, followed by swap proposals between a subset of
(neighboring) chains constitutes a single PT iteration. It is easy to see that
repeating this process will cause fast-mixing samples to eventually be swapped
into the lower temperatures, thus (from the point of view of $p_1$) performing
a long-distance MCMC move. In this manner, the samples of $p_1$ can move much
faster across configurations and more accurately reflect the model statistics. 

\paragraph{Deep Belief Networks.} RBMs are typically trained and stacked in a
greedy layer-wise fashion to form a Deep Belief Network (DBN). The result is a
model which learns a joint distribution over multiple layers of latent
variables \footnote{Formally, a DBN with $3$ latent layers defines a
joint-distribution $p(v,h_1,h_2,h_3) = p_1(v\mid h_1) p_2(h_1 \mid h_2) p_3(h_2
, h_3)$, where $p_i$ is the distribution of the $i$-th level}. These
can then be used as input to a classifier, most often outperforming features
extracted by shallow models (i.e. a single RBM). Another interesting property
of DBNs however, is that the last level RBM is known to mix well across its
input configurations. On MNIST, \cite{Hinton06} showed that drawing samples
from the top-level RBM via Gibbs sampling and then deterministically projecting
these back to the input space, led to samples which mixed very well across
digit classes. This important result would appear to be somewhat
under-appreciated, especially considering the mixing issues typically associated with single
layer models, even when used in conjunction with tempering.

This phenomenon has received renewed attention of late.
\cite{Luo+al-AISTATS2013-small} have found that depth is crucial in sampling
and learning better texture models in the RBM family. The relationship between
depth and mixing was further explored in \cite{Bengio-et-al-ICML2013}, where
the authors hypothesize that good mixing is the result of deep networks
learning to disentangle the underlying factors of variation.

\section{Deep Tempering}
\label{sec:deeptempering}

Our proposed solution starts with the idea of {\it learning} the set of
interpolating distributions. A possible realization of this idea would be to
simulate an ensemble $\prod_{i=1}^M p_i(v; \theta_i)$, with each distribution $p_i$ having
its own set of parameters $\theta_i$ and $|\theta_1| > |\theta_2| >
\dots > |\theta_M|$.  Each $p_i$ would be trained independently to model the
empirical distribution $q(v)$, but regularized so that $p_i$'s become easier
to sample from \footnote{This could be implemented in many different
forms, e.g. by increasingly constraining the weight-norms with $i$} with
index $i$. By construction, $KL(p_i, p_{i+1})$ would be kept small and should
thus allow for frequent replica exchange. While such an algorithm would
certainly lower the computational cost of simulating ``higher temperatures'',
it does little to address the large number of interpolating distributions often
required by tempering methods.

Exploiting the observation that good mixing can be achieved by remapping
the data into a high-level latent space, we instead \emph{learn} interpolating distributions $p_i$
via stacked latent variable models and train each $p_i$ to model the posterior
distribution of $p_{i-1}$. Training this ensemble jointly, we can then
propose replica exchanges between neighboring models and exploit samples of
the upper layers to perform long distance MCMC moves in the visible space
of lower layers. While this procedure is generally applicable to the
learning of any energy-based model featuring latent variables, this paper
considers the particular case where $p_i$ is a Restricted Boltzmann Machine
(RBM) \cite{Smolensky86}.

Similar to Parallel Tempering, Deep Tempering simulates an ensemble $\mathcal{P}$ of models defined
by Equation~\ref{eq:dt_ensemble}.
The key difference is that models $p_i$ do not share the same
parametrization.
Each
$p_i$ defines a joint distribution over its own set of visible variables $v_i
\in \Omega_i$ and latent variables $h_i \in \Psi_i$.
The support of these variables is constrained such that $\Omega_{i} =
\Psi_{i-1}$. The parameters $\theta_i$ of these distributions are learned
jointly, in order for $p_i$ to accurately reflect the posterior distribution of
the lower layer.  Defining $q_0:=q(x)$ to be the empirical distribution and
$q_i(x)$, the $i$-th layer aggregate posterior distribution defined as
$q_{i>0}(x) := \mathbb{E}_{v_i \sim q_{i-1}} \left[ p_i (h_i \mid v_i) \right]$,
parameters of the $i$-th layer are trained simultaneously, e.g. through maximum likelihood,
as shown in Equation~\ref{eq:dt_learning}.
\begin{align}
    \label{eq:dt_ensemble}
    \mathcal{P} = \prod_{i=1}^M \left\{ \sum_{h_i} p_i(v_i, h_i; \theta_i) \right\}. \\
    \label{eq:dt_learning}
    \theta_i^* := \text{argmax}\ 
    \mathbb{E}_{q_{i-1}}
        \left[ \log \sum_{h_i}\ p_{i} \left(v_{i}, h_{i} \right) \right].
\end{align}
The particular form of proposal models $p_i$ is left free, with the only
constraint that $p_i(v_i)$ and $p_i(h_i)$ can be evaluated up to a
normalization constant, meaning to say $v_i$ and $h_i$ can each be marginalized
analytically. This last fact allows us to perform cross-temperature state
swaps between neighboring distributions. Replica exchanges can then be considered
between $\tilde{h}_i \sim p_i(h_i)$ and $\tilde{v}_{i+1} \sim
p_{i+1}(v_{i+1})$, with a swap acceptance probability $r_i$ given by:
\begin{align}
    \label{eq:dt_swap}
    r_{i}
    &= min\left( 1, \frac
        {
            p_i \left( h_i = \tilde{v}_{i+1} \right)
            p_{i+1} \left( v_{i+1} = \tilde{h}_{i} \right)
        }
        {
            p_i \left( h_i = \tilde{h}_i \right)
            p_{i+1} \left( v_{i+1} = \tilde{v}_{i+1} \right)
        }
        \right).
\end{align}

The Deep Tempering method as a whole is detailed in the algorithm of Figure~\ref{fig:alg}.

\begin{figure}
\centering
\begin{minipage}{.49\textwidth}
  \centering
    \begin{algorithm}[H]
    \caption{${\tt DTLearn(q, \{\theta_i, \tilde{v}_i^-, isodd \}}$}
    \begin{algorithmic}
    \vspace{4mm}
    \STATE {\small $\{\tilde{v}_i\} \leftarrow DTNegSwaps(\{ \theta_i, \tilde{v}_1^-, isodd \})$ }
    \FOR{$i = 1:M$}
        \STATE {\small $\tilde{v}_i^+ \sim q\ $ if $(i==0)\ $ else
                $\ \mathbb{E}_{q_{i-1}} \left[ p_i(h_i \mid v_i) \right]$}
        \STATE {\small $\Delta\theta_i \leftarrow {\tt SMLGrad}(\theta_i, \tilde{v}_i^+, \tilde{v}_i^-).$}
        \STATE {\small $\theta_i \leftarrow \theta_i + \Delta\theta_i$.}
    \ENDFOR
    \STATE {\small $isodd \leftarrow$ {\bf not} $isodd$.}
    \vspace{4mm}
    \end{algorithmic}
    \end{algorithm}
\end{minipage}
\begin{minipage}{.49\textwidth}
  \centering
    \begin{algorithm}[H]
    \caption{${\tt DTNegSwaps(\{\theta_i, \tilde{v}_i, isodd \}}$}
    \begin{algorithmic}
    \FOR{ {\small $i \in$ $1$:$M$:$2$ if $\ isodd\ $ else $\ 2$:$M$:$2$} }
    \STATE {\tiny $s_i = -F_i(\tilde{v}_{i+1}) - F_{i+1}(\tilde{h}_i) + F_i(\tilde{h}_{i+1}) + F_{i+1}(\tilde{v}_{i+1})$}
    \STATE {\small Swap $\tilde{h}_i$ and $\tilde{v}_{i+1}$ if $rand() < r_i := \exp s_i$.}
    \STATE {\small $\tilde{v}_i \sim p_i(v \mid h = \tilde{h}_i)$.}
    \ENDFOR
    \FOR{$i = 1:M$}
    \STATE 
    {\small
        $\tilde{h}_i \sim p_i(h_i \mid v=\tilde{v}_i);\ \ $
        $\tilde{v}_i \sim p_i(v_i \mid h=\tilde{h}_i)$}
    \ENDFOR
    \STATE {\small Return $\{ \tilde{v}_i \}$.}
    \end{algorithmic}
    \end{algorithm}
\end{minipage}
\captionof{figure}{
{\small {\bf Deep Tempering algorithm}. ${\tt DTLearn}$ implements a
single step of gradient descent on the empirical distribution $q$.
${\tt DTNegSwaps}$ starts by generating samples $\{ \tilde{v}_i^-\}$ from the
ensemble $\mathcal{P}$, by proposing state swaps between neighboring models,
followed by Gibbs sampling at each layer.
``Positive phase'' samples $\tilde{v}_i^+, \forall i \in [1,M]$ are then
generated according to the typical DBN greedy inference process.  Given
samples $\tilde{v}_i^+$, $\tilde{v}_i^-$, ${\tt SMLGrad}$ (not shown) computes
the typical SML gradient independently for each layer. Note that the above
obfuscates the fact that in practice, $v_i^+$ and $v_i^-$ are typically
mini-batches of samples.} \label{fig:alg}}
\end{figure}

While our proposed method shares many similarities to parallel tempering,
there are nonetheless important differences which will have consequences
for training RBMs. As previously described, parallel tempering exploits a
temperature parameter to be able to simulate samples from a
high-temperature (i.e. smoothed-out) version of the  
model distribution. In our case, the analogous
high-temperature distribution is not only simulating samples in a
transformed space (that corresponding to the RBM latent variables), but the
upper layer model is actually trained not to reflect the model
distribution, but rather a transformation of the data distribution given by
the RBM conditional $p(h \mid v)$. Despite these differences, accept-reject
step still ensures that the samples accepted to swap into the lower-layer
RBM do properly reflect the lower-layer RBM model distribution

\paragraph{Deep Belief Network Joint-Training.}
Viewing the proposed approach as simply a means of training the lower-layer
RBM ignores one important point: at the end of the training process, we
effectively have not a single trained RBM, but a stack of RBMs trained
together as a whole deep belief network (DBN). From this perspective, the
proposed approach can be seen as a means of joint-training a DBN. 
This form of training is in stark contrast to the common practice of
greedily training the constituent RBMs, in a bottom-up sequential manner.

In the proposed approach, joint training plays a
crucial role in ensuring that at every point in training, upper-layer RBMs
faithfully represent the distribution over
the lower layer hidden units in order to ensure adequate acceptance rate of
the state-swapping MCMC moves.
From this perspective, if our goal is to learn a DBN, the proposed approach
is doubly efficient. Not only is the learned-tempering model also a viable DBN in
its own right, but we are able to exploit the rapid mixing of the upper-layer
RBMs to ensure rapid (and consistent) mixing of the RBMs at every-layer of the
DBN.
 
It is worthwhile to ask if this style of joint-training 
results in any important differences compared to the traditional greedy layer-wise
training strategy. We will return to this question in the experimental
section. For now it suffices to point to some interesting and potentially
important consequences for the proposed style of joint training.
First, without swapping between neighboring RBMs, the proposed approach is
simply the simultaneous application of the standard layer-wise RBM training
procedure with one notable exception. The upper-layer RBMs are effectively
training to reach a moving target: the distribution that they are being
asked to learn is constantly changing as the lower-layer RBMs learn.  
Second, each RBM is trained using SML where the negative phase samples are
drawn using MCMC over an ensemble of models, mimicking the effect of
traditional tempered MCMC chains. Typically, this chain will
consist of small jumps that reflect the local structure of the RBM around
negative phase samples, and long distance jumps that correspond to particle
swaps between the states of neighboring RBMs. 
The fact that all RBMs share both the positive phase data samples as well
as the negative phase samples will help maintain consistency between the models,
ensuring acceptable trans-RBM swap ratios. 

As RBM training proceeds and the energy surface begins to reflect the
distribution of the data, the low-level RBMs will increasingly rely on the
upper-layer RBMs to provide -- through swaps -- a steady supply of diverse
particles for use in the negative phase of training. This is similar to the
situation when RBMs are trained with a tempering strategy. However unlike
in true tempering, here the upper-layer RBMs do not share the same
parametrization as the lower-layer RBM. As a result, replica exchanges may not
help escape ``spurious'' local minimas in the energy which may emerge as a
result of training (and which are not shared across layers). Instead, we expect the
long distance MCMC moves offered by DT to be biased towards modes which are
well supported by the data.

\section{Experiments}
\label{sec:experiments}

\begin{figure}
    \subfloat[RBM]
    {
        \label{fig:modes1a}
        \includegraphics[scale=0.32]{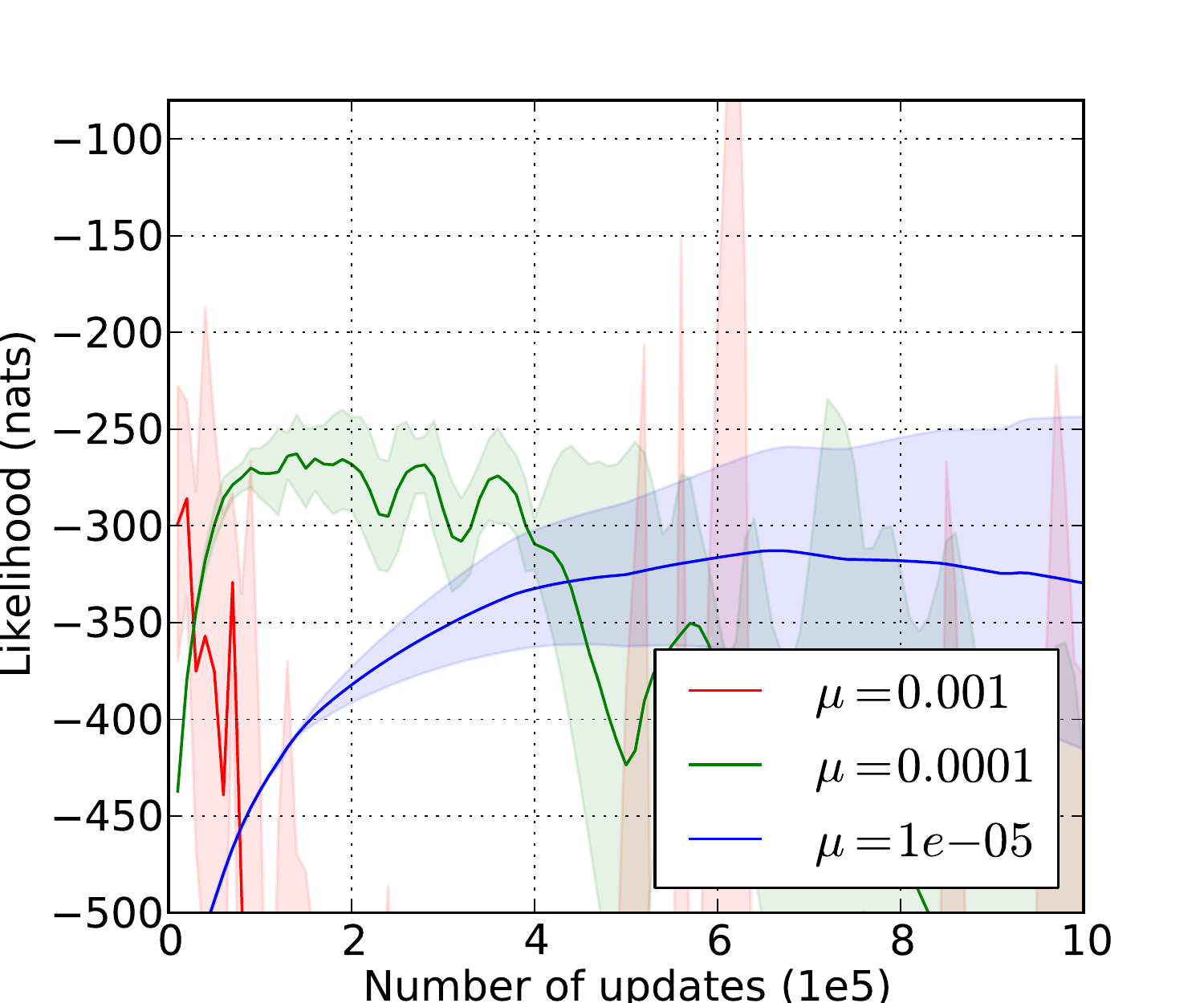}
    }
    \subfloat[tDBN-2]
    {
        \label{fig:modes1b}
        \includegraphics[scale=0.32]{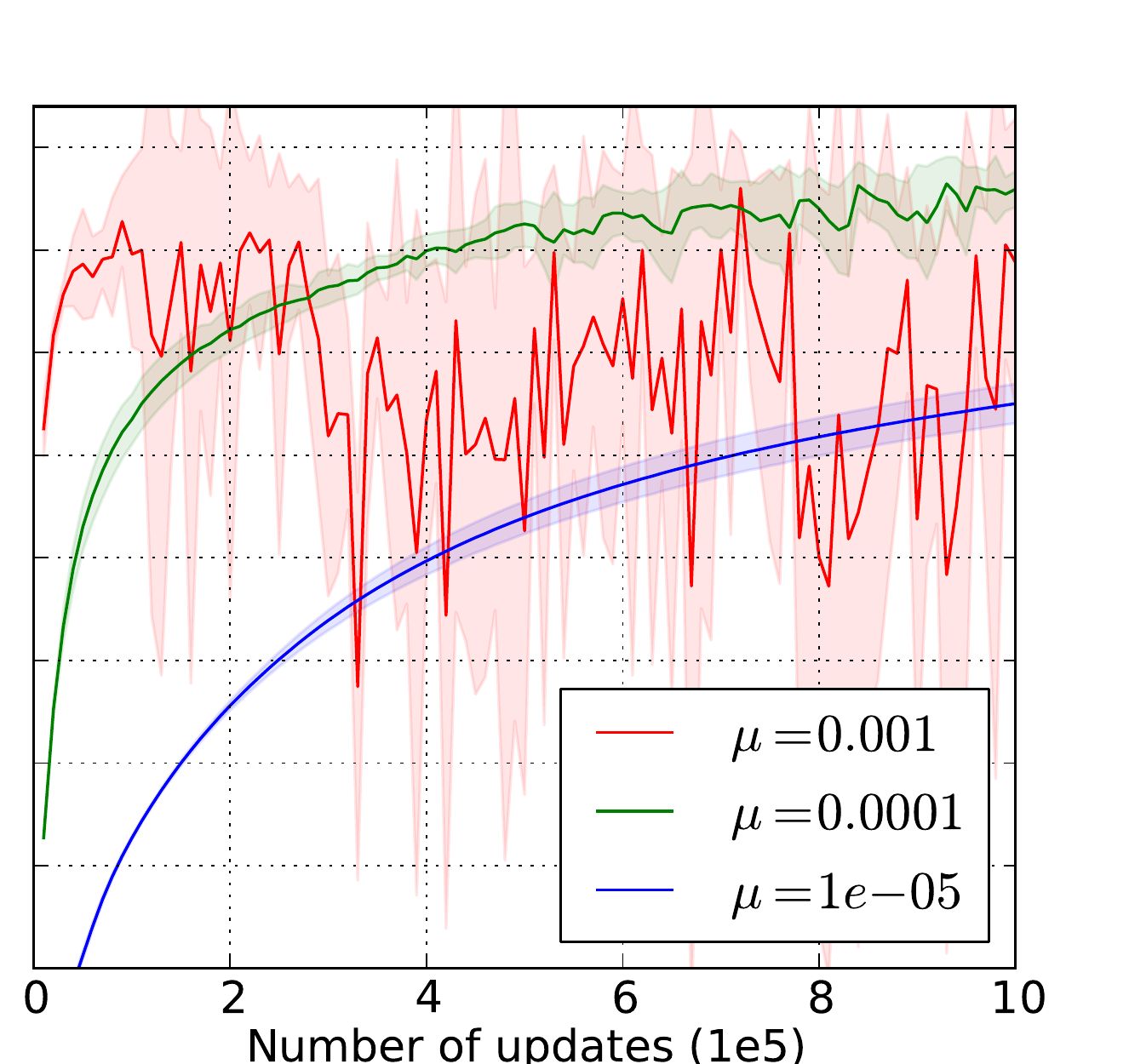}
    }
    \subfloat[tDBN-3]
    {
        \label{fig:modes1c}
        \includegraphics[scale=0.32]{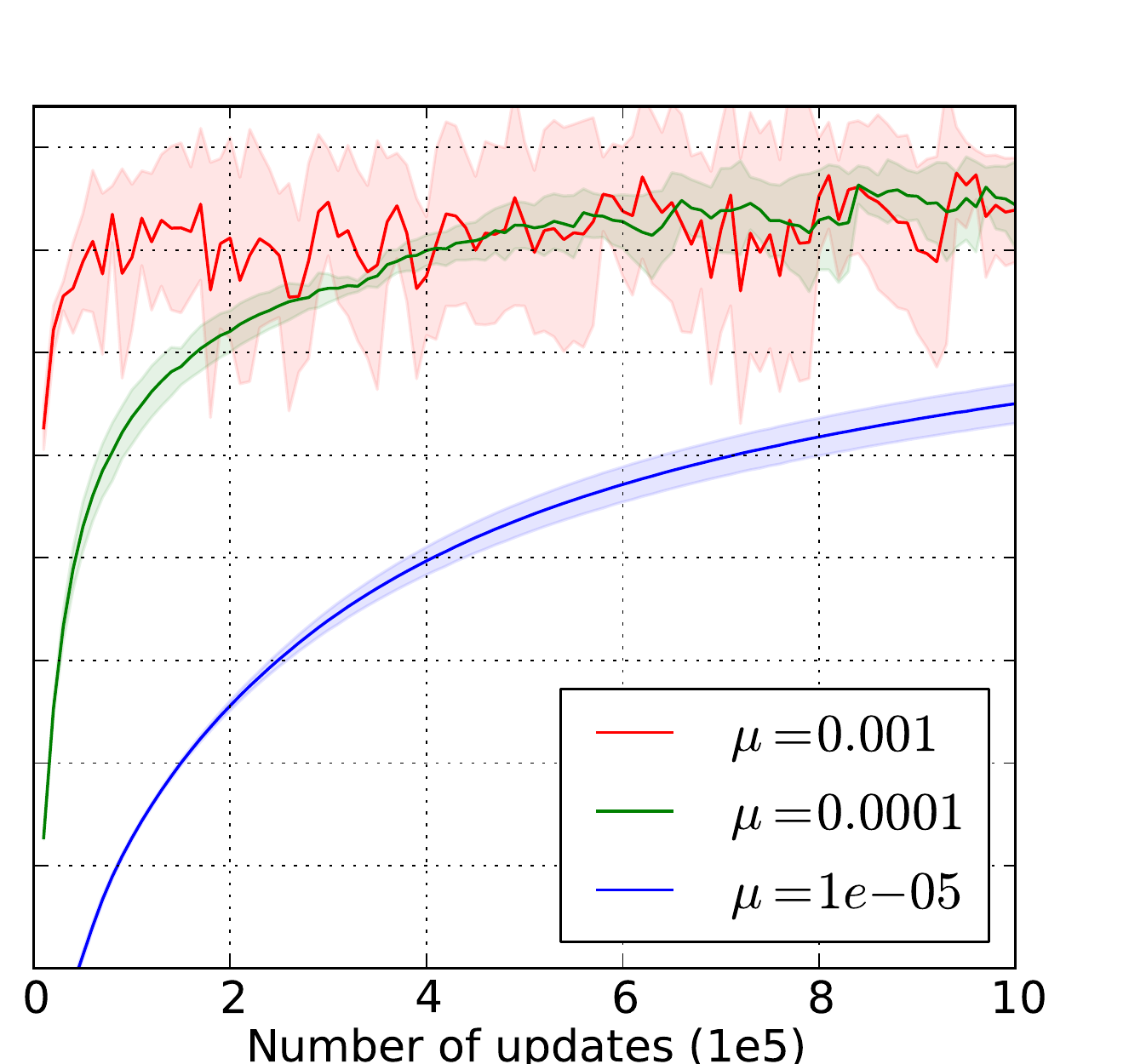}
    }
    \caption[]{
    {\small Sensitivity analysis of the test set likelihood as a function of the learning rate $\mu$.
    Models considered are (a) a single layer RBM model (b) a 2-layer t-DBN and
    (c) a 3-layer t-DBN. Shaded areas represent $\pm$ one standard deviation.}
    \label{fig:modes1}}
\end{figure}

We start our evaluation of Deep Tempering on the ``Artificial
Modes'' dataset first introduced in \cite{Desjardins-wkshp-2010}. The dataset
is characterized by five basic $28\times 28$ binary random noise images.
Training data is generated online as a mixture model over these five modes,
permuting each pixel independently with a mode dependent probability.
This dataset is difficult from a sampling perspective, as it combines modes of
low probability having large spatial support, and narrow modes with high
probability. On the other hand, the structure of the data is relatively simple,
allowing us to use models with low capacity and thus measure likelihood exactly
during training. Likelihood is reported on a test set of $5$k images.

\paragraph{RBM vs. t-DBN} Figure~\ref{fig:modes1} compares the test-set
likelihood curve (during training) of a $10$-hidden unit RBM, vs the likelihood
of the first-level RBM in a 2 and 3-layer tempered DBN (t-DBN), with $10$
hidden units per layer. We therefore evaluate DT as a better training/sampling
algorithm for RBMs.  Gradients were computed by averaging over a small
mini-batch of size $5$, kept small so as to prevent models from covering all
modes through a large mini-batch of (effectively) stationary particles. To
simplify the analysis and given the interesting dynamic between learning and
mixing (the mixing rate of a sampler constrains the set of feasible learning
rates), we focus our hyper-parameter search to constant learning rates in
$\{10^{-3},10^{-4},10^{-5}\}$. Results are averaged over $5$ seeds, with the
shaded area representing $\pm \sigma$.

As we can see, the single RBM is unable to learn a meaningful model of the
dataset for any setting of the learning rate. SML diverges when using high
learning rates and yields an unimpressive score of around $-329.6$ nats when
using a learning rate of $10^{-5}$. In contrast, a stack of two RBMs trained
via Deep Tempering achieves a score of $-120.4$ nats with a learning rate of
$10^{-4}$. Pushing the learning rate further to $10^{-3}$ causes the two layer
model to diverge, a fact which seems attenuated by moving to a stack of 3 RBMs,
shown in Figure~\ref{fig:modes1c}. Cross-model state swaps occurred rather
frequently on this dataset, with the best performing t-DBN2 model swapping
$32\%$ of the time on average.

\begin{figure}
    \subfloat
    {
        \label{fig:modes_cast}
        \includegraphics[scale=0.35]{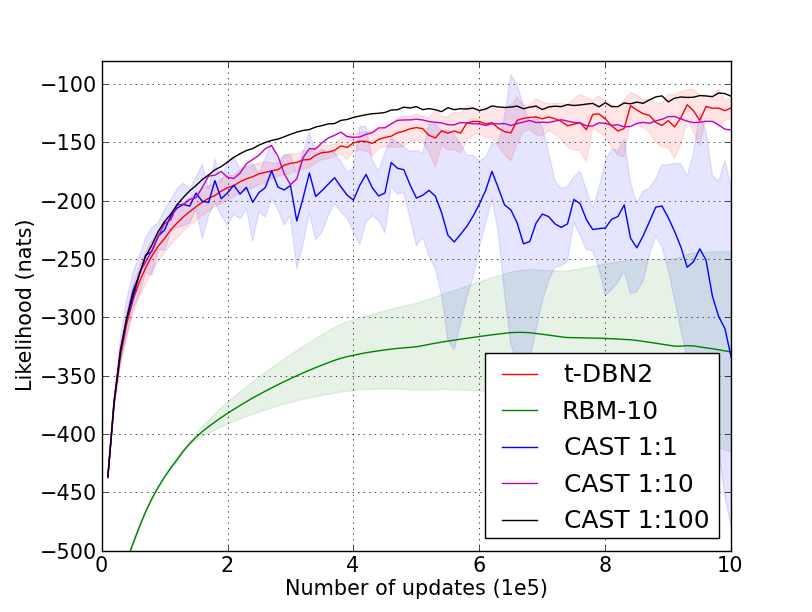}
    }
    \subfloat
    {
        \label{fig:modes_dbn}
        \includegraphics[scale=0.35]{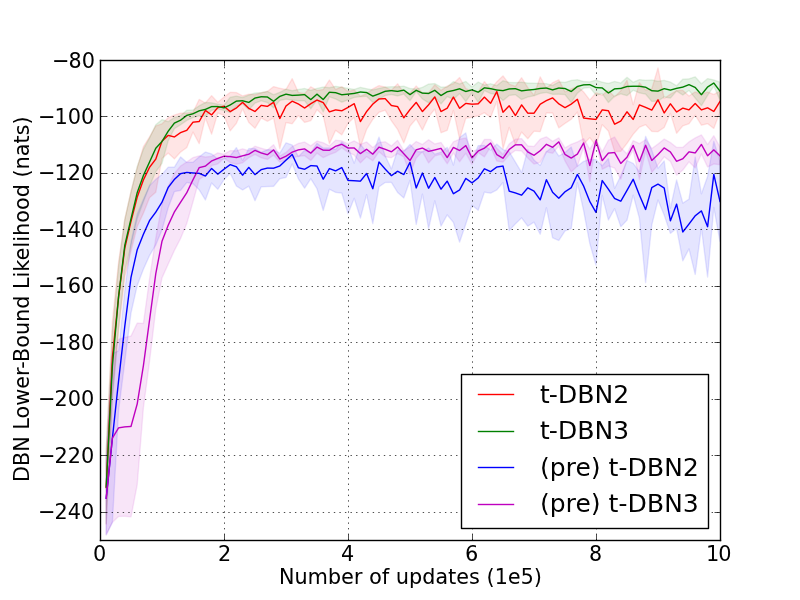}
    }
    \caption[]{
    {\small {\bf (left)} Likelihood of the first layer RBM for various model
    configurations / training algorithms. RBM-10 was selected as the best
    single-layer model from those of Figure~\ref{fig:modes1a}, t-DBN2 from
    Figures~(\ref{fig:modes1b}, \ref{fig:modes1c}). CAST 1:X shows the likelihood
    of an RBM trained with Coupled Adaptive Simulated Tempering, having a ratio
    of non-tempered to tempered chains of 1/X.
    {\bf (right)} DBN likelihood lower-bound for the best 2 and 3 layer t-DBNs
    (joint-trained only, 2 top curves) vs. t-DBNs having first been pre-trained before
    the joint-training procedure (2 bottom curves).}}
\end{figure}

\paragraph{PT, CAST vs t-DBN.} DT also compares favorably to state-of-the-art tempering methods. In
\cite{Desjardins-wkshp-2010}, Parallel Tempering required around $50$ tempered
chains spaced uniformly in the unit interval, to achieve a likelihood of around
$-175$ nats. Optimizing the inverse temperature parameters through an adaptive
tempering mechanism improved this likelihood to around $-150$ nats, while requiring
fewer tempered chains ($20$).
Results obtained with the original CAST formulation \cite{Salakhutdinov-ICML2010}
(CAST 1:1), using $100$ tempered chains spaced uniformly in the unit interval,
are shown in Figure~\ref{fig:modes_cast}. The model originally performs well
but eventually diverges despite AST converging to the proper (log) adaptive
weights (given by the log-partition functions at each temperature). Suspecting
insufficient replica exchanges between tempered and non-tempered chains (given
the large number of temperatures being simulated), we also trained a one layer
RBM using a modified version of CAST, where we increase the ``coupling ratio'',
i.e. the ratio of non-tempered to tempered chains, denoted as CAST 1:X.
A ratio of 1:10 led to notable improvements in likelihood, achieving $-139.3$
nats after $10^6$ updates. CAST 1:100 achieved a new state-of-the-art for
single-layer models: $-110.5$ nats. The reader should keep in mind however
that CAST $1:100$ with a mini-batch of size $5$ requires simulating $505$
Markov chains in total, compared to the $10$ Markov chains required by a
t-DBN2.

\paragraph{DBN Lower-Bound.}
As mentioned in Section~\ref{sec:deeptempering}, DT applied to a stack of RBMs
can equivalently be viewed as performing joint-training of a DBN. It is thus
interesting to study the effect of DT on the lower-bound of the DBN likelihood,
as formulated in \citep{Salakhutdinov+Murray-2008}. The results
are shown in Figure~\ref{fig:modes_dbn}. The optimal learning rate achieved a
lower-bound of $-94.6$ nats for a 2-layer DBN (shown in green), and $-91.0$
nats for the 3-layer variant (red). One may also wonder about the effect of
joint-training vs. the typical greedy layer-wise training procedure employed by
DBNs. To avoid the extreme instability of SML trained RBMs on this dataset, we
adjusted the greedy layer-wise training to incorporate early-stopping, based on
detecting oscillations in training likelihood. $M-1$ layers were pre-trained in
this manner, followed by joint-training across all layers. The results clearly
show worse performance for the pre-trained models. This suggests that the greedy
layer-wise pre-training scheme of DBNs leads to poor local minima.

\section{Discussion}
\label{sec:discussion}

Deep Tempering combines ideas from traditional MCMC tempering methods, with new
observations from the field of deep learning. From the field of MCMC, we borrow
the idea of simulating an ensemble of distributions, where the fast-mixing
properties of high-temperature chains is leveraged to facilitate training of
the lower layers, which are often prone to getting stuck in regions of high
probability. From the world of deep learning, we exploit recent findings which
show that models at deeper layers in the hierarchy seem to naturally exhibit
good mixing. These two ideas are combined through a joint-training strategy,
which considers individual models in a deep hierarchy as interpolating distributions
for the lower layers.
While we have concentrated on the application of Deep Tempering to RBMs
the method is much more widely applicable. Perhaps most interesting is that
it can be readily applied to Deep Boltzmann Machines (DBM) \cite{SalHinton09}
both to promote mixing in the negative phase of learning as well as to draw
diverse samples from a fully trained model. We simply construct a latent
variable model over all the even layers of the DBM and use this model to
propose swaps with the DBM state over the even layers of the model.

The method that we describe in this paper can be considered an example of
an auxiliary variable method Markov Chain Monte Carlo sampling of the lower
layer RBM. Similar to other such methods, including the celebrated Sweden-Wang method
\cite{Swendsen1987}, one can interpret the addition of the random variables
in the upper-layer RBMs (both visibles and hiddens) as auxiliary variables
that are added to augment the simulated system with the goal of achieving
more efficient sampling. Where the proposed approach differs from more
standard auxiliary variable methods for MCMC is the richness in structure
relating these auxiliary variables both to themselves and to the target
lower-layer RBM. While Swendsen-Wang can be thought of as imposing a
clustering on the variables of the simulated system, our approach uses a
DBN (with $M-1$ layers) to model the dependency structures between the
latent variables of the lower-layer RBM. 

\small{
\bibliography{strings,strings-shorter,ml,aigaion-shorter}
\bibliographystyle{natbib}
}

\end{document}